%% file: main.tex
\documentclass[a4paper,UKenglish,cleveref, autoref, thm-restate]{lipics-v2021}

\hideLIPIcs

\title{Efficient Certified Reasoning for Binarized Neural Networks} %
\titlerunning{Efficient Certified Reasoning for Binarized Neural Networks} %

\author{Jiong Yang}{Georgia Institute of Technology, United States}{}{}{}
\author{Yong Kiam Tan}{Institute for Infocomm Research (I$^2$R), A*STAR, Singapore \and Nanyang Technological University, Singapore}{yongkiam.tan@ntu.edu.sg}{https://orcid.org/0000-0001-7033-2463}{Singapore NRF Fellowship Programme NRF-NRFF16-2024-0002}
\author{Mate Soos}{University of Toronto, Canada}{}{}{}
\author{Magnus O. Myreen}{Chalmers University of Technology, Sweden \and University of Gothenburg, Sweden}{myreen@chalmers.se}{https://orcid.org/0000-0002-9504-4107}{Swedish Research Council grant 2021-05165}
\author{Kuldeep S. Meel}{Georgia Institute of Technology, United States \and  University of Toronto, Canada}{}{}{}

\authorrunning{J. Yang et al.} %

\Copyright{Jiong Yang, Yong Kiam Tan, Mate Soos, Magnus O. Myreen, and Kuldeep S. Meel} %



\category{} %

\relatedversion{} %

\newcommand{\cmsbnn}{\ensuremath{\mathsf{CryptoMiniSat\mbox{-}BNN}}}
\newcommand{\fratxorbnn}{\ensuremath{\mathsf{FRAT\mbox{-}xor\mbox{-}bnn}}}
\newcommand{\cakexlrup}{\ensuremath{\mathsf{cake\underline{\hspace{0.4em}}xlrup\mbox{-}BNN}}}
\newcommand{\ApproxMCBNN}{\ensuremath{\mathsf{ApproxMC\mbox{-}BNN}}}
\newcommand{\ApproxMCCertBNN}{\ensuremath{\mathsf{ApproxMCCert\mbox{-}BNN}}}
\newcommand{\CertCheckBNN}{\ensuremath{\mathsf{CertCheck\mbox{-}BNN}}}
\supplement{We have integrated our implementation into the following existing open-source repositories.}
\supplementdetails[{subcategory={\cmsbnn}}]{Software}{https://github.com/msoos/cryptominisat}
\supplementdetails[{subcategory={{\fratxorbnn} and {\cakexlrup}}}]{Software}{https://github.com/meelgroup/frat-xor}
\supplementdetails[{subcategory={\ApproxMCBNN}}]{Software}{https://github.com/meelgroup/approxmc}
\supplementdetails[{subcategory={{\ApproxMCCertBNN} and {\CertCheckBNN}}}]{Software}{https://github.com/meelgroup/approxmc-cert}
\supplementdetails[{subcategory={Proof Format}}]{Text}{https://github.com/meelgroup/frat-xor/blob/main/format.md}
\supplementdetails[{subcategory={Benchmark}}]{Dataset}{https://doi.org/10.5281/zenodo.15630195}

\funding{This project was supported in part by Natural Sciences and Engineering Research Council of Canada (NSERC), funding reference [RGPIN-2024-05956].
}
\acknowledgements{We sincerely thank the anonymous reviewers for their constructive feedback on our earlier draft of this paper. The computational work for this article was performed on resources of the National Supercomputing Centre, Singapore https://www.nscc.sg, and Niagara supercomputer at the SciNet HPC Consortium. SciNet is funded by Innovation, Science and Economic Development Canada; the Digital Research Alliance of Canada; the Ontario Research Fund: Research Excellence; and the University of Toronto.}%

\nolinenumbers %

\usepackage{booktabs}
\hypersetup{
	colorlinks=true,
	linkcolor=blue,
	filecolor=magenta,
	urlcolor=blue,
	citecolor = blue,
}
\usepackage{adjustbox}
\usepackage{amsmath}
\usepackage{amsfonts}
\usepackage{algorithm}
\usepackage[noend]{algpseudocode}
\usepackage{mathtools}
\usepackage{fancyvrb}
\usepackage{xcolor}

\newcommand{\bigO}[1]{\ensuremath{\mathcal{O}\left({#1}\right)}}

\newcommand{\ApproxMC}{\ensuremath{\mathsf{ApproxMC}}}
\newcommand{\ApproxMCPB}{\ensuremath{\mathsf{ApproxMC\mbox{-}PB}}}
\newcommand{\appmcbnn}{\ApproxMCBNN}

\newcommand{\ApproxMCCert}{\ensuremath{\mathsf{ApproxMCCert}}}
\newcommand{\CertCheck}{\ensuremath{\mathsf{CertCheck}}}
\newcommand{\Ganak}{\ensuremath{\mathsf{Ganak}}}
\newcommand{\PBCount}{\ensuremath{\mathsf{PBCount}}}
\newcommand{\arjun}{\ensuremath{\mathsf{Arjun}}}
\newcommand{\cms}{\ensuremath{\mathsf{CryptoMiniSat}}}

\newcommand{\roundingsat}{\ensuremath{\mathsf{RoundingSat}}}
\newcommand{\minisatcs}{\ensuremath{\mathsf{MiniSatCS}}}
\newcommand{\cadical}{\ensuremath{\mathsf{CaDiCaL}}}
\newcommand{\veripb}{\ensuremath{\mathsf{VeriPB}}}
\newcommand{\cakelpr}{\ensuremath{\mathsf{cake\underline{\hspace{0.4em}}lpr}}}
\newcommand{\true}{\texttt{True}}
\newcommand{\false}{\texttt{False}}

\newcommand{\qhead}{\ensuremath{\mathsf{qhead}}}
\newcommand{\trail}{\ensuremath{\mathsf{trail}}}
\newcommand{\getsize}{\ensuremath{\mathsf{GetSize}}}
\newcommand{\getwatch}{\ensuremath{\mathsf{GetWatches}}}
\newcommand{\watches}{\ensuremath{\mathsf{watches}}}
\newcommand{\confl}{\ensuremath{\mathsf{conflict}}}
\newcommand{\propcl}{\ensuremath{\mathsf{PropagateClause}}}
\newcommand{\propbnn}{\ensuremath{\mathsf{PropagateBnn}}}
\newcommand{\prop}{\ensuremath{\mathsf{Propagate}}}
\newcommand{\gje}{\ensuremath{\mathsf{GaussJordanElim}}}
\newcommand{\tcnt}{\ensuremath{\mathsf{trueCount}}}
\newcommand{\gettcnt}{\ensuremath{\mathsf{GetTrueCount}}}
\newcommand{\getlcnt}{\ensuremath{\mathsf{GetLiteralCount}}}
\newcommand{\ucnt}{\ensuremath{\mathsf{undefCount}}}
\newcommand{\getucnt}{\ensuremath{\mathsf{GetUndefCount}}}
\newcommand{\getcutoff}{\ensuremath{\mathsf{GetCutOff}}}
\newcommand{\getout}{\ensuremath{\mathsf{GetOutputLiteral}}}
\newcommand{\minisat}{\ensuremath{\mathsf{MiniSat}}}

\newcommand{\appmc}{\ApproxMC}

\newcommand{\isabellehol}{\ensuremath{\mathsf{Isabelle/HOL}}}
\newcommand{\Prob}[1]{\ensuremath{\mathsf{Pr}\left[{#1}\right]}}
\newcommand{\sol}{\ensuremath{\mathtt{sol}(\varphi)}}
\newcommand{\xor}{\text{XOR}}
\newcommand{\bnn}{\text{BNN}}

\newcommand{\fratxor}{\ensuremath{\mathsf{FRAT\mbox{-}xor}}}

\definecolor{highlightred}{rgb}{.8, 0.0, 0.0}
\definecolor{highlightgreen}{HTML}{427E3B}
\definecolor{highlightyellow}{HTML}{EE9700}
\newcommand{\greenc}[1]{\textcolor{highlightgreen}{#1}}
\newcommand{\yellowc}[1]{\textcolor{highlightyellow}{#1}}

\newcommand{\sign}{\ensuremath{\mathsf{sign}}}
\newcommand{\abcrown}{\ensuremath{\mathsf{\alpha,\beta\mbox{-}CROWN}}}
\newcommand{\cora}{\ensuremath{\mathsf{CORA}}}
\newcommand{\nnv}{\ensuremath{\mathsf{NNV}}}
\newcommand{\marabou}{\ensuremath{\mathsf{Marabou}}}
\newcommand{\neuralsat}{\ensuremath{\mathsf{NeuralSAT}}}
\newcommand{\reluplex}{\ensuremath{\mathsf{Reluplex}}}
\newcommand{\pyrat}{\ensuremath{\mathsf{PyRAT}}}
\newcommand{\nnenum}{\ensuremath{\mathsf{nnenum}}}
\newcommand{\nevertwo}{\ensuremath{\mathsf{NeVer2}}}
\newcommand{\gurobi}{\ensuremath{\mathsf{Gurobi}}}
\newcommand{\matlab}{\ensuremath{\mathsf{MATLAB}}}

\begin{document}

\maketitle

\input{sec/abstract}
\input{sec/introduction}

\input{sec/preliminary}

\input{sec/related_work}

\input{sec/theory}

\input{sec/experiment}
\input{sec/conclusion}

\appendix

\input{sec/appendix}

\bibliography{main}

\end{document}

%% file: sec/abstract.tex
\begin{abstract}
	Neural networks have emerged as essential components in safety-critical applications---these use cases demand complex, yet trustworthy computations.
	Binarized Neural Networks (BNNs) are a type of neural network where each neuron is constrained to a Boolean value;
  they are particularly well-suited for safety-critical tasks because they retain much of the computational capacities of full-scale (floating-point or quantized) deep neural networks, but remain compatible with satisfiability solvers for qualitative verification and with model counters for quantitative reasoning.
	However, existing methods for BNN analysis suffer from either limited scalability or susceptibility to soundness errors, which hinders their applicability in real-world scenarios.

	In this work, we present a scalable and trustworthy approach for both qualitative and quantitative verification of BNNs.
	Our approach introduces a native representation of BNN constraints in a custom-designed solver for qualitative reasoning, and in an approximate model counter for quantitative reasoning.
	We further develop specialized proof generation and checking pipelines with native support for BNN constraint reasoning, ensuring trustworthiness for all of our verification results.
	Empirical evaluations on a BNN robustness verification benchmark suite demonstrate that our certified solving approach achieves a $9\times$ speedup over prior certified CNF and PB-based approaches, and our certified counting approach achieves a $218\times$ speedup over the existing CNF-based baseline.
	In terms of coverage, our pipeline produces fully certified results for $99\%$ and $86\%$ of the qualitative and quantitative reasoning queries on BNNs, respectively.
  This is in sharp contrast to the best existing baselines which can fully certify only $62\%$ and $4\%$ of the queries, respectively.
\end{abstract}

%% file: sec/introduction.tex
\section{Introduction} \label{sec:intro}
Neural networks have had a transformative impact on fields such as image recognition~\cite{KSH12}, natural language processing~\cite{DCLT19}, and decision making~\cite{SHMG+16}, achieving remarkable success in tackling complex tasks in each of those domains.
This success has led to their potential deployment in critical scenarios, such as autonomous driving~\cite{BDDF+16}, aircraft collision avoidance~\cite{JLBO+16}, and drug discovery~\cite{VKVT+15}.
However, despite the impressive performance of neural networks, they often exhibit unexpected behaviors and their lack of explainability makes them difficult to control~\cite{SZSB+14}; this has led to significant concerns about their use in high-stakes applications.

Verification techniques for neural networks can help address some of these safety and security concerns~\cite{BILV+16,HKWW17,KBDJ+17,WZXL+21,WIZT+24}.
Here, we focus on verification for \emph{binarized neural networks} (BNNs)~\cite{HCSE+16,JR20,NKRS+18}, i.e., where the input/output of every neuron is quantized to a single bit.
Our focus is motivated by two practical reasons.
First, BNNs are computationally attractive for real-world, resource-constrained use cases---the absence of floating-point arithmetic in BNNs eliminates floating-point issues, and the bit-level representation enables faster inference speeds and more compact memory layouts~\cite{HLZL+23,HCSE+16,MNSM+17,RORF+16,ZGCL+23}.
Second, existing studies have demonstrated that BNNs can be verified against quantitative specifications, which is beyond the reach of verification methods for more general classes of neural networks~\cite{BSSM+19}.

To elaborate on the latter point, the Boolean nature of BNNs allows both the networks and their specifications to be encoded as Boolean formulas which, in turn, enables the use of off-the-shelf SAT solvers and model counters to verify those specifications.
For example, prior research has explored both qualitative and quantitative reasoning for BNN robustness, susceptibility to Trojan attacks, and fairness specifications~\cite{BSSM+19,JR20,NKRS+18,YM21}.
Unfortunately, existing combinatorial solving methods for such analyses suffer from limited scalability---they apply only to toy-sized BNNs with hundreds of neurons and a few layers.
In contrast, custom methods based on abstraction and bounds propagation for verifying input-output specifications scale to much larger (and more general) neural networks~\cite{KBDJ+17,WZXL+21,WIZT+24}, but are susceptible to errors; in the annual neural network verification competitions, participating tools have been repeatedly shown to produce incorrect conclusions.
\emph{Certified reasoning}~\cite{MMNS11}, where a tool generates both a conclusion and an independently-checkable proof of that conclusion, has long become a mainstay of trustworthiness for modern SAT solvers~\cite{HHKW17,P17,DBLP:journals/sttt/TanHM23,WHH14} but it remains nascent for neural network verification tools~\cite{WIZT+24}.

In this work, we address the challenges of scalability and trustworthiness in BNN verification by developing an efficient and certified approach for both qualitative and quantitative reasoning on BNNs.
Our contributions are as follows.
\begin{itemize}
\item For qualitative reasoning, we integrate a native representation of BNN constraints into a modern CNF-XOR SAT solver to enhance its BNN solving efficiency.
\item We then extend the solver's associated UNSAT proof format and verified proof checker with native support for BNN constraints, together enabling an efficient solving and certification pipeline for CNF-XOR-BNN formulas.
\item Building upon this pipeline, we further develop a certified approximate model counter for quantitative reasoning over BNNs, leveraging native XOR and BNN representations for both effective reasoning and certification.
\end{itemize}

Empirically, our approach achieves state-of-the-art certified performance in both qualitative and quantitative reasoning for BNNs.
Notably, we observe that the compactness of proof certificates with native BNN proof steps enables fast, verified proof checking.
\begin{itemize}
	\item For qualitative reasoning, our end-to-end approach produced certified answers for $99\%$ of the benchmark UNSAT queries, achieving a $9\times$ speedup over alternative CNF- and PB-based approaches.
	\item For quantitative reasoning, our certified counting approach answered $86\%$ of the queries, with $218\times$ speedup over the baseline which could only fully certify $4\%$ of the queries.
\end{itemize}

By developing solving and certification in tandem, our work offers a trustworthy approach to BNN verification with promising scalability.
In fact, during the development of our certification pipeline, we identified and fixed a bug in our solver's implementation of the watching scheme for BNN constraints, which highlights the practical importance of certification. %

The rest of this paper is organized as follows.
Section~\ref{sec:prelim} presents the preliminaries on BNNs used throughout the paper.
Section~\ref{sec:related-work} discusses further background and related work.
Section~\ref{sec:theory} introduces the proposed certified solving and counting approaches for BNN verification, and Section~\ref{sec:exp} evaluates their empirical performance against existing methods.
We summarize our findings in Section~\ref{sec:conclusion}.

%% file: sec/preliminary.tex
\section{Preliminaries: Binarized Neural Networks} \label{sec:prelim}

Given a set of Boolean variables $\{x_1, x_2, \ldots, x_n\}$, a \emph{literal} $l$ is either a variable $x$ or its negation $\neg x$. 
A \emph{clause} $C$ is a disjunction of $k > 0$ literals: $C \coloneqq l_1 \vee l_2 \vee \ldots \vee l_k$.
A \emph{Conjunctive Normal Form (CNF) formula} $\varphi$ is a conjunction of $m > 0$ clauses: $\varphi \coloneqq C_1 \wedge C_2 \wedge \ldots \wedge C_m$.
A \emph{solution} to $\varphi$ is an assignment of truth values to the variables such that $\varphi$ evaluates to $\true$.
The formula $\varphi$ is \emph{satisfiable} if at least one solution exists, and \emph{unsatisfiable} otherwise.
The \emph{model count} of $\varphi$ refers to the total number of solutions.

Binarized Neural Networks (BNNs) are a class of neural networks in which each neuron is constrained to a Boolean value.
Given an input tensor $\mathbf{x} \in \{0, 1\}^n$, an output tensor $\mathbf{y} \in \{0, 1\}^m$, a weight matrix $\mathbf{w} \in \{-1,1\}^{n\times m}$, and a bias vector $\mathbf{b}\in \mathbb{Z}^m$, a BNN layer maps $\mathbf{x}$ to $\mathbf{y}$ as follows:
\begin{align*}
	\mathbf{y} = \sign\left(\mathbf{w}^\top \mathbf{x} + \mathbf{b}\right)
\end{align*}
Here, the {\sign} function represents the non-linear sign activation used in BNNs, functioning as a binarized analogue of the ReLU activation.
A fully connected BNN consists of several layers laid out in sequence, i.e., the output of one layer is the input of the next, and so on; they can be trained using the straight-through estimator, which enables gradient flow through the sign function during backpropagation.
For additional details on BNN training and architecture, we refer the readers to~\cite{HCSE+16}.

A BNN constraint corresponds to a neuron in the network, encoding the logical relationship between its input neurons and its output.
Formally, we define this relationship as follows:
\begin{definition}[BNN Constraint]
	\label{def:bnn}
	Let $y$ denote the output of a given neuron, and let $x_1, x_2, \ldots, x_n$ denote the values of its $n$ input neurons. Let $w_i$ represent the weight associated with the connection from the $i$-th input neuron, and $b$ denote the bias term. Then, the logical input-output relationship of a BNN constraint is defined as:
	\begin{align}
	\label{eq:bnn}
		y \leftrightarrow \sum_{i=1}^{n} w_i x_i + b \ge 0
	\end{align}
	where $y$ and $x_i$ are Boolean variables; $w_i$ and $b$ are constants with $w_i \in \{1, -1\}$ for $i \in [1..n]$ and $b \in \mathbb{Z}$.
\end{definition}

We model the value of each binarized neuron as a Boolean variable (in contrast to $y, x_i \in \mathbb{R}$ in conventional neural networks).
Some prior work represents neuron values using $\{-1, 1\}$; the two notations are interchangeable via a simple linear transformation.
In this work, we adopt the Boolean representation for clarity and compatibility with SAT-based reasoning.
The greater-than-or-equal operator in Equation~\ref{eq:bnn} encodes the {\sign} function.

A BNN constraint can be readily transformed into an equivalent \emph{conditional cardinality constraint}, which is defined below. In the rest of this paper, we use the terms BNN constraint and conditional cardinality constraint interchangeably when the context is clear.

\begin{definition}[Conditional Cardinality Constraint]
	\label{def:cond-card}
	Given literals $l_1, l_2, \ldots, l_n$, an output literal $l_y$, and an integer constant $k$, a conditional cardinality constraint is defined as:
	\begin{align}
		\label{eq:cond-card}
		l_y \leftrightarrow \sum_{i=1}^{n} l_i \ge k
	\end{align}
	where $l_y$ represents the truth value of the cardinality constraint $\sum_{i=1}^{n} l_i \ge k$, 
	and $l_1, l_2, \ldots, l_n$ are referred as the left-hand-side (LHS) literals.
\end{definition}

Using the above constraint representations, a BNN with $t$ layers, input bits $\mathbf{l}^0$, and output bits $\mathbf{l}^{t}$ can be represented by a Boolean formula $\text{BNN}(\mathbf{l}^0,\mathbf{l}^{t})$.
An input-output specification of the BNN is then a Boolean formula $\text{Prop}\left(\mathbf{l}^0, \mathbf{l}^t\right)$.
For instance, a common safety specification is the robustness property, which assesses either the existence (qualitative reasoning) or the number (quantitative reasoning) of adversarial inputs.
This property is defined as follows:
\begin{definition}[BNN Robustness]
  A BNN is $\varepsilon$-robust for the input $\mathbf{l}^0_g$ and output
  $\mathbf{l}^t_g$ if there are no satisfying assignments for the following
  property; $\|\cdot\|_1$ denotes the Hamming distance.
	\begin{align*}
		\textnormal{Prop}\left(\mathbf{l}^0, \mathbf{l}^t\right) \coloneqq \| \mathbf{l}^0 - \mathbf{l}^0_g \|_1 \le \varepsilon \ \wedge \ \mathbf{l}^t \not= \mathbf{l}^t_g
	\end{align*}
\end{definition}
The Boolean nature of BNNs and their property specifications makes it possible to leverage existing solvers and model counters for answering both qualitative and quantitative robustness queries~\cite{BSSM+19,JR20,NKRS+18,YM21}.
However, due to the inherent statistical nature of neural network training, the existence of adversarial inputs is expected, making purely qualitative queries potentially uninformative.
Quantitatively estimating the prevalence of adversarial inputs allows us to draw statistically significant conclusions~\cite{BSSM+19}.
Moreover, the absence of floating-point arithmetic in the formulation of BNNs eliminates floating-point issues when answering these critical robustness queries.

%% file: sec/related_work.tex
\section{Related Work} \label{sec:related-work}

\subparagraph{Neural Network Verification.}
Neural network verification has received growing attention, as reflected in the organization of annual benchmark competitions~\cite{BBJW24}.
Existing techniques span several categories, including linear bound propagation (e.g., {\abcrown}~\cite{WZXL+21}), reachability analysis (e.g., {\cora}~\cite{A15}, \nnv~\cite{LCTJ23}, {\nnenum}~\cite{B21}, {\nevertwo}~\cite{DGPT24}, and {\pyrat}~\cite{LLG24}), and SMT-based solving (e.g., {\marabou}~\cite{WIZT+24}, {\neuralsat}~\cite{DXND24}, and {\reluplex}~\cite{KBDJ+17}).
These approaches focus on qualitative reasoning, particularly the robustness of ReLU networks~\cite{GBB11}, and often rely on commercial solvers such as {\gurobi} or {\matlab}.
However, no fully verified proof checker currently exists for traditional neural network verification frameworks~\cite{DIPS+23}.
Another line of research targets the verification of BNNs.
This includes qualitative reasoning for robustness~\cite{JR20,NKRS+18}, quantitative reasoning for robustness, susceptibility to Trojan attacks, and fairness specifications~\cite{BSSM+19,YM21}, efforts to accelerate verification through increased sparsity in BNNs~\cite{JR20,NZGW20}, and techniques for networks with both binarized and non-binarized neurons~\cite{DBLP:conf/tacas/Amir0BK21}.

\subparagraph{BNN Encodings.}
With the conditional cardinality constraint representation, BNN constraints can be further encoded into clauses via cardinality encodings or represented directly as PB constraints.
Encoding BNNs into CNF formulas results in a substantial increase in formula size, primarily due to the large number of conditional cardinality constraints.
For a BNN with $m$ neurons, where each neuron is represented using Equation~\ref{eq:bnn} with $n$ Boolean variables, the CNF encoding introduces $\bigO{n}$ variables and clauses per neuron, leading to a total of $\bigO{nm}$ variables and clauses.
As a result, the formula size can easily reach millions of variables and clauses even for networks with only a few hundred neurons, which poses significant challenges for existing SAT solvers and model counters.
The complete CNF encoding of BNNs is presented in~\cite{BSSM+19,NKRS+18}, where BNNs are first encoded as conditional cardinality constraints and subsequently translated into clauses using cardinality encodings. A PB-based encoding for BNNs can be found in~\cite{YM21}.

\subparagraph{Propagation and Conflict Analysis for BNN Constraints}
To avoid the cumbersome clause-based encoding of Equation~\ref{eq:bnn}, Jia and Rinard introduced an alternative representation of BNN constraints in their solver, {\minisatcs}, which enables efficient detection of propagation and conflicts specific to BNNs~\cite{JR20}. {\minisatcs} transforms Equation~\ref{eq:bnn} into a reified cardinality constraint, which is logically equivalent to the conditional cardinality constraint in Equation~\ref{eq:cond-card}.
Propagation and conflict detection over Equation~\ref{eq:cond-card} are handled as follows.
\begin{itemize}
	\item \textbf{Operand-inferring:} If $l_y$ is assigned {\true} and $n-k$ literals among $l_1, \ldots, l_n$ are already assigned {\false}, then the remaining unassigned $l_i$ must be assigned {\true}; otherwise, a conflict is detected.
	Conversely, if $l_y$ is assigned {\false} and $k-1$ literals among $l_1, \ldots, l_n$ are assigned {\true}, then the remaining unassigned literals must be assigned {\false}.
	\item \textbf{Target-inferring:} If $k$ literals among $l_1, \ldots, l_n$  are assigned {\true}, then $l_y$ must be assigned {\true}; otherwise, a conflict is detected.
	Similarly, if $n-k+1$ literals among $l_1, \ldots, l_n$ are assigned {\false}, then $l_y$ must be assigned {\false}.
\end{itemize}

When a conflict is detected, the responsible literals are added to the conflict clause.
This design eliminates the need to encode Equation~\ref{eq:bnn} into large sets of clauses, significantly reducing formula size.
The native support for direct propagation and conflict detection allows for faster solving and scalability to networks with thousands of neurons.
However, reified cardinality constraints are not supported by the UNSAT proof format used in standard SAT solvers, rendering the results uncertifiable without further developments.

\subparagraph{Certified Solving and Counting.}
Certification has become a cornerstone of trustworthiness in the SAT community~\cite{HHKW17,P17,DBLP:journals/sttt/TanHM23,WHH14}.
Recent efforts have extended certification to the CNF-XOR solver, {\cms} and even to probabilistic counting~\cite{TYSM+24}.
The model counting problem aims to determine the number of satisfying assignments for a Boolean formula,
while \emph{approximate model counting} seeks to compute a high-quality approximation of this count; approximate model counters have shown practical performance in many applications~\cite{BSSM+19,DMPV17,GVF22}.
The state-of-the-art algorithm, {\ApproxMC}~\cite{CMV13,YM23}, provides a $(1+\varepsilon)$ approximation of the model count with confidence at least $1-\delta$, given tolerance $\varepsilon$ and confidence $\delta$.
To ensure correctness, a certification pipeline has been developed for {\ApproxMC}~\cite{TYSM+24}, enabling independent verification of its output.
In parallel, recent work has also begun to explore \emph{certified exact model counting}~\cite{BNAH23,FHR22}.

%% file: sec/theory.tex
\section{Certified Reasoning for BNNs} \label{sec:theory}
This section presents our integration of native BNN constraints for BNN solving, counting, and proof checking.
Section~\ref{sec:qualitative-reasoning} introduces our new solver, {\cmsbnn}, which incorporates native support for BNN constraints to enable efficient qualitative reasoning on BNNs; it also describes our proof format with native support for BNN reasoning in unsatisfiability proofs.
Section~\ref{sec:quantitative-reasoning} describes our certified model counter, {\ApproxMCCertBNN}, which similarly leverages native BNN constraint support for efficient and certified quantitative reasoning.
Finally, Section~\ref{sec:bnn-bug-report} concludes with a case study of a subtle bug in the watching scheme of {\cmsbnn}, uncovered by our certification pipeline---this highlights the complementary role of certification in ensuring the correctness of automated reasoning tools.

\subsection{Certified Solving for Qualitative BNN Reasoning}
\label{sec:qualitative-reasoning}
Our new solver efficiently reasons over both XOR and BNN constraints, collectively referred to as CNF-XOR-BNN formulas;
support for XOR constraints is essential for CNF-BNN model counting (which will be presented in Section~\ref{sec:quantitative-reasoning}).
We formally define CNF-XOR-BNN formulas as follows:
\begin{definition}[CNF-XOR-BNN Formula]
	Given a set of clauses $C_1, C_2, \ldots, C_m$, XOR constraints $\xor_1, \xor_2, \ldots, \xor_t$, and BNN constraints $\bnn_1, \bnn_2, \ldots, \bnn_s$,
	a CNF-XOR-BNN formula $\varphi$ is defined as the conjunction of these constraints:
	\begin{align*}
		\varphi \coloneqq \bigwedge_{i=1}^m C_i \wedge \bigwedge_{j=1}^t \xor_j \wedge \bigwedge_{k=1}^s \bnn_k
	\end{align*}
A solution (or model) $\omega$ for $\varphi$ is a Boolean assignment to variables that simultaneously satisfies all of the clauses, XORs, and BNN constraints.
\end{definition}

\subsubsection{Solving CNF-XOR-BNN Formulas} \label{sec:bnn-solving}

\subparagraph{Input format.}
Figure~\ref{fig:cnf-xor-bnn} (left) shows an example of a CNF-XOR-BNN formula in our input format, which extends the standard DIMACS format~\cite{JT96} to support XOR and BNN constraints.
In the header, the first integer specifies the number of variables, and the second indicates the total number of constraints, including clauses, XORs, and BNNs.
Each line beginning with the prefix \textbf{b} represents a BNN constraint, while lines starting with the prefix \textbf{x} denote XOR constraints.
The last line in Figure~\ref{fig:cnf-xor-bnn} (left) encodes a BNN constraint corresponding to $x_1 + \neg x_2 + x_3 \ge 2 \leftrightarrow x_4$.
The list of integers following the prefix \textbf{b}, terminated by a $0$, specifies the LHS literals of the cardinality constraint, in this case $x_1, \neg x_2$, and $x_3$.
The last two integers, following the terminating $0$, indicate the cutoff value and the output literal, respectively.
In this example, $2$ is the cutoff value, and $4$ denotes the output literal $x_4$.
Similarly, the line \texttt{x 1 -2 -3 0} encodes an XOR constraint: $x_1 \oplus \neg x_2 \oplus \neg x_3 = 1$.

\begin{figure}
	\centering
	\begin{minipage}[t]{0.3\textwidth}
	\centering
	{CNF-XOR-BNN formula}
	\begin{Verbatim}[fontsize=\small,commandchars=\\\{\}]
p cnf 4 5
1 -2 0
-1 3 0
\greenc{x} 1 -2 -3 0
-4 0
\greenc{b} 1 -2 3 0 2 4 0
	\end{Verbatim}
	\end{minipage}
	\quad
	\begin{minipage}[t]{0.3\textwidth}
		\centering
		{FRAT-XOR-BNN proof}
		\begin{Verbatim}[fontsize=\small,commandchars=\\\{\}]
\greenc{o} \yellowc{1} 1 -2 0
\greenc{o} \yellowc{2} -1 3 0
\greenc{o x} \yellowc{1} 1 -2 -3 0
\greenc{o} \yellowc{3} -4 0
\greenc{o b} \yellowc{1} 1 -2 3 0 \greenc{k} 2 4 0
\greenc{i} \yellowc{4} -1 -3 0 \greenc{b l} \yellowc{1} 0 \greenc{u} \yellowc{3} 0
\greenc{i} \yellowc{5} 2 -3 0 \greenc{b l} \yellowc{1} 0 \greenc{u} \yellowc{3} 0
\greenc{i} \yellowc{6} -3 0 \greenc{l} \yellowc{1 5 4} 0
\greenc{a} \yellowc{7} -1 0
\greenc{a} \yellowc{8} -2 0
\greenc{i} \yellowc{9} 1 2 3 0 \greenc{l} \yellowc{1} 0
\greenc{a} \yellowc{10} 0
\greenc{f} \yellowc{1} 1 -2 0
\greenc{f} ...
\greenc{f} \yellowc{10} 0
\greenc{f x} \yellowc{1} 1 -2 -3 0
\greenc{f b} \yellowc{1} 1 -2 3 0 \greenc{k} 2 4 0
		\end{Verbatim}
	\end{minipage}
	\quad
	\begin{minipage}[t]{0.3\textwidth}
		\centering
		{XLRUP proof}
		\begin{Verbatim}[fontsize=\small,commandchars=\\\{\}]
\greenc{o x} \yellowc{1} 1 -2 -3 0
\greenc{i cb} \yellowc{4} -1 -3 0 \yellowc{1} \greenc{u} \yellowc{3} 0
\greenc{i cb} \yellowc{5} 2 -3 0 \yellowc{1} \greenc{u} \yellowc{3} 0
\yellowc{5} \greenc{d} \yellowc{3} 0
\yellowc{6} -3 0 \yellowc{4 1 5} 0
\yellowc{6} \greenc{d} \yellowc{5 4} 0
\yellowc{7} -1 0 \yellowc{6 2} 0
\yellowc{7} \greenc{d} \yellowc{2} 0
\yellowc{8} -2 0 \yellowc{7 1} 0
\yellowc{8} \greenc{d} \yellowc{1} 0
\greenc{i cx} \yellowc{9} 1 2 3 0 \yellowc{1} 0
\yellowc{10} 0 \yellowc{7 6 9 8} 0

		\end{Verbatim}
	\end{minipage}
	\caption{
		Example of a CNF-XOR-BNN formula (left) and its unsatisfiability proof in FRAT-XOR-BNN (middle) and XLRUP (right) formats. Green highlights indicate special keywords, while yellow highlights denote constraint indices.}
	\label{fig:cnf-xor-bnn}
\end{figure}

\subparagraph{CNF-XOR-BNN Solving.}
A CNF-XOR-BNN solver takes a formula $\varphi$ in the extended DIMACS format and determines its satisfiability.
We introduce the first CNF-XOR-BNN solver, $\cmsbnn$, by extending an existing CNF-XOR solver, $\cms$~\cite{DBLP:conf/sat/SoosNC09}, with native support for BNN constraints.
Our implementation maintains an internal representation of each BNN constraint in the form of a conditional cardinality constraint and directly detects when it causes a unit propagation or conflict.
The solver follows the standard Conflict-Driven Clause Learning (CDCL) procedure used in modern SAT solvers while incorporating specialized propagation and conflict detection schemes for XOR and BNN constraints, as outlined in Algorithms~\ref{alg:prop} and~\ref{alg:prop-bnn}.

In Algorithm~\ref{alg:prop}, the unit propagation procedure iteratively processes unit literals stored in $\trail$, starting from the index $\qhead$.
For each unit literal $l$, the solver retrieves its watched constraints in Line~\ref{ln:prop-get-watch}, which may correspond to either a clause or a BNN constraint; XOR watches are maintained separately.
The procedure then iterates over all watched constraints in Lines~\ref{ln:prop-loop-watch-begin}--\ref{ln:prop-loop-watch-end}.
If a clause is watched, clausal unit propagation is performed in Line~\ref{ln:prop-clause-prop}.
If a BNN constraint is watched, a specialized BNN propagation procedure, {\propbnn} (Algorithm~\ref{alg:prop-bnn}), is executed in Line~\ref{ln:prop-bnn-prop}.
If no conflict is detected during clausal and BNN unit propagation, Gauss-Jordan elimination is applied to propagate the literal $l$ over XOR constraints and to detect potential conflicts (Line~\ref{ln:prop-gje}).
Finally, if no conflict is encountered, the index $\qhead$ is incremented and the loop continues.

\begin{algorithm}[t]
	\caption{$\prop(\qhead, \trail)$}
	\label{alg:prop}
	\begin{algorithmic}[1]
		\State $\confl \gets$ empty clause
		\While{$\qhead < \getsize(\trail)$ and $\confl$ is empty}
			\State $l \gets \trail[\qhead]$
			\State $\watches \gets \getwatch[\neg l]$ \label{ln:prop-get-watch}
			\For{$w \in \watches$} \label{ln:prop-loop-watch-begin}
				\If{$w$ is a clause}
					\State $\confl \gets \propcl(w, l)$ \label{ln:prop-clause-prop}
				\Else
					\State $\confl \gets \propbnn(w, l)$ \label{ln:prop-bnn-prop}
				\EndIf
			\EndFor \label{ln:prop-loop-watch-end}
			\If{$\confl$ is empty}
				\State $\confl \gets \gje(l)$ \label{ln:prop-gje}
			\EndIf
			\State $\qhead \gets \qhead + 1$
		\EndWhile
		\State \Return $\confl$
	\end{algorithmic}
\end{algorithm}

Algorithm~\ref{alg:prop-bnn} describes the procedure, $\propbnn(w, l)$, for unit propagation and conflict clause generation from a BNN constraint $w$ with respect to a literal $l$.
Let $n$ and $k$ denote the number of LHS literals and the cutoff value of the cardinality constraint in $w$, respectively, and let $l_y$ denote the output literal.
For each BNN constraint, the solver maintains two counters: the number of literals assigned {\true} ({\tcnt}) and the number of unassigned literals ({\ucnt}), to enable prompt detection of unit propagation or conflicts.
If $l$ is a positive LHS literal in $w$, {\tcnt} is incremented and {\ucnt} is decremented (Lines~\ref{ln:prop-bnn-l-pos-begin}--\ref{ln:prop-bnn-l-pos-end});
if $l$ is a negative LHS literal, {\ucnt} is decremented (Line~\ref{ln:prop-bnn-l-neg}).
Lines~\ref{ln:prop-bnn-prop-begin}--\ref{ln:prop-bnn-prop-end} perform unit propagation in a BNN constraint based on the values of {\tcnt} and {\ucnt}.
Specifically, Lines~\ref{ln:prop-bnn-prop-begin}--\ref{ln:prop-bnn-op-infer-end} handle the propagation of LHS literals.
If $l_y$ is {\true} and the sum of {\tcnt} and {\ucnt} is less than $k$ (Line~\ref{ln:prop-bnn-yt-lt-k}),
a conflict is detected because the cardinality constraint cannot be satisfied even if all unassigned LHS literals are assigned {\true}, contradicting the assignment $l_y=\true$.
This situation arises when at least $n-k+1$ LHS literals are assigned {\false}, causing the sum of {\tcnt} and {\ucnt} to fall below $k$.
Consequently, $\neg l_y$ and the $n-k+1$ {\false} LHS literals are added to the conflict clause (Line~\ref{ln:prop-bnn-yt-confl}).
Alternatively, if the sum of {\tcnt} and {\ucnt} equals $k$, all unassigned LHS literals must be assigned {\true} (Line~\ref{ln:prop-bnn-yt-prop}) to satisfy the assignment $l_y=\true$.
Similarly, conflict detection and assignment inference for the case where $l_y$ is assigned {\false} are handled in Lines~\ref{ln:prop-bnn-yf-begin}--\ref{ln:prop-bnn-op-infer-end}.
On the other hand, Lines~\ref{ln:prop-bnn-out-infer-begin}--\ref{ln:prop-bnn-prop-end} handle the propagation of the output literal.
If $\tcnt \ge k$, meaning that at least $k$ LHS literals are assigned {\true}, the cardinality constraint in $w$ is satisfied, and the output literal $l_y$ must be assigned {\true} (Line~\ref{ln:prop-bnn-out-t}).
Alternatively, if $\tcnt + \ucnt < k$, it is impossible to satisfy the cardinality constraint, and $l_y$ must be assigned {\false} (Line~\ref{ln:prop-bnn-out-f}).

Whenever a BNN constraint infers the assignment of a literal, a reason clause is generated to facilitate conflict analysis in the CDCL algorithm.
For instance, when $l_y$ is inferred to be {\true} in Line~\ref{ln:prop-bnn-out-t}, the reason clause includes $l_y$ and the negation of the $k$ LHS literals assigned to {\true}, thereby explaining the inference of $l_y$ based on these assignments.

\begin{algorithm}[t]
	\caption{$\propbnn(w, l)$}
	\label{alg:prop-bnn}
	\begin{algorithmic}[1]
		\State $n \gets \getlcnt(w)$
		\State $k \gets \getcutoff(w)$
		\State $l_y \gets \getout(w)$
		\State $\tcnt \gets \gettcnt(w)$
		\State $\ucnt \gets \getucnt(w)$
		\If{$l$ is a positive literal in $w$}
			\State $\tcnt = \tcnt + 1$ \label{ln:prop-bnn-l-pos-begin}
			\State $\ucnt = \ucnt - 1$ \label{ln:prop-bnn-l-pos-end}
		\ElsIf{$l$ is a negative literal in $w$}
				\State $\ucnt = \ucnt - 1$ \label{ln:prop-bnn-l-neg}
		\EndIf
		\If{$l_y$ is $\true$} \label{ln:prop-bnn-prop-begin}
			\If{$\tcnt + \ucnt < k$} \label{ln:prop-bnn-yt-lt-k}
				\State add $\neg l_y$ and $n-k+1$ {\false} literals to $\confl$. \label{ln:prop-bnn-yt-confl}
			 \ElsIf{$\tcnt + \ucnt == k}$
			 	\State assign {\true} to all unassigned literals. \label{ln:prop-bnn-yt-prop}
			 \EndIf
		\ElsIf{$l_y$ is $\false$} \label{ln:prop-bnn-yf-begin}
			\If{$\tcnt \ge k$}
				\State add $l_y$ and the negations of $k$ {\true} literals to $\confl$.
			\ElsIf{$\tcnt + 1 == k$}
				\State assign {\false} to all unassigned literals. \label{ln:prop-bnn-yf-prop}
			\EndIf \label{ln:prop-bnn-op-infer-end}
		\ElsIf{$\tcnt \ge k$} \label{ln:prop-bnn-out-infer-begin}
			\State assign {\true} to $l_y$. \label{ln:prop-bnn-out-t}
		\ElsIf{$\tcnt + \ucnt < k$}
			\State assign {\false} to $l_y$. \label{ln:prop-bnn-out-f}
		\EndIf \label{ln:prop-bnn-prop-end}
	\end{algorithmic}
\end{algorithm}

\subparagraph{Comparison.} {\cmsbnn} differs from {\minisatcs}~\cite{JR20} in several key aspects.
First, we maintain BNN constraints in the standard AtLeastK form, as defined in Equation~\ref{eq:cond-card}, whereas {\minisatcs} represents them as AtMostK cardinality constraints.
Second, our solver generates minimal conflict and reason clause from BNN constraints, while {\minisatcs} does not.
For instance, when $l_y$ is assigned {\false} and at least $k$ LHS literals are assigned {\true}, {\minisatcs} includes the negation of all {\true} LHS literals in the conflict clause, in addition to $l_y$.
In contrast, {\cmsbnn} adds only the negations of exactly $k$ {\true} literals, which constitute the minimal reason for the conflict.
Third, {\cmsbnn} adopts a unified data structure to consistently maintain watches for both clauses and BNN constraints, whereas {\minisatcs} employs separate data structures for clause and BNN watches.
Lastly, {\minisatcs} is built on top of {\minisat}, whereas {\cmsbnn} integrates the BNN constraints directly in the state-of-the-art CNF-XOR solver, {\cms}, which incorporates more advanced techniques than {\minisat}.
The native support for both XOR and BNN constraints in {\cmsbnn} is crucial for our development of a certified CNF-BNN approximate model counter in Section~\ref{sec:quantitative-reasoning}.

\subsubsection{Certifying CNF-XOR-BNN Solving} \label{sec:bnn-solving-cert}
Our new pipeline for certified CNF-XOR-BNN unsatisfiability consists of several parts.
\begin{enumerate}
	\item We extend the FRAT-XOR\footnote{The FRAT-XOR format is itself an extension of the FRAT format~\cite{BCH21} with support for XOR-related reasoning steps.} format~\cite{TYSM+24} by introducing support for BNN-specific reasoning steps, resulting in an extended FRAT-XOR-BNN format that can be readily generated by {\cmsbnn}.
	\item We develop a format elaborator, {\fratxorbnn}, which converts FRAT-XOR-BNN proofs into an elaborated proof format, XLRUP (eXtended LRUP). This translator extends {\fratxor}~\cite{TYSM+24}, originally designed for FRAT-XOR proofs.
	\item The XLRUP proof format supports RUP proofs, extended with both XOR- and BNN-specific steps; we design, implement, and formally verify an efficient proof checker for XLRUP proofs (specifically, we add new BNN reasoning support).
\end{enumerate}

Continuing the example from Figure~\ref{fig:cnf-xor-bnn}, the unsatisfiability proof for its CNF-XOR-BNN formula in both FRAT-XOR-BNN and XLRUP formats is displayed (middle and right columns, respectively); green highlights indicate special keywords, while yellow highlights denote the constraint indices.

\subparagraph{FRAT-XOR-BNN format.}
In the FRAT-XOR-BNN proof format, BNN-related steps are marked with the keyword \textbf{b}.
The prefix \textbf{o b} denotes an original BNN constraint in the input formula, while the prefix \textbf{f b} indicates a final BNN constraint that remains after the empty clause is derived.
The \texttt{clause-from-bnn} step is prefixed with the identifier \textbf{i} (denoting implication), followed by the clause implied by a BNN constraint.
The keyword \textbf{b l} specifies the index of the BNN constraint, while the keyword \textbf{u} lists the set of unit clauses used to simply the BNN constraint.

For example, the following step records the derivation of a new clause $(\neg x_1 \vee \neg x_3)$ to be stored at index 4:
\begin{Verbatim}[fontsize=\small,commandchars=\\\{\}]
	\greenc{i} \yellowc{4} -1 -3 0 \greenc{b l} \yellowc{1} 0 \greenc{u} \yellowc{3} 0
\end{Verbatim}
This clause is implied by the first BNN constraint:
\begin{align}
	\label{eq:bnn-eg}
	x_1 + \neg x_2 + x_3 \ge 2 \leftrightarrow x_4
\end{align}
together with the unit clause $\neg x_4$ at index 3.
The chain of unit clauses simplifies the BNN constraint by applying the corresponding variable assignments.

Similarly, the \texttt{clause-from-xor} step is prefixed with the identifier \textbf{i}, followed by the keyword \textbf{l}, which indicates the XOR constraints implying the clause.
For instance, the following step records the derivation of a new clause $(x_2 \vee x_1 \vee x_3)$ with index 9:
\begin{Verbatim}[fontsize=\small,commandchars=\\\{\}]
	\greenc{i} \yellowc{9} 1 2 3 0 \greenc{l} \yellowc{1} 0
\end{Verbatim}
This clause is implied by the first XOR constraint:
\begin{align}
	\label{eq:xor-eg}
	x_1 \oplus \neg x_2 \oplus \neg x_3 = 1
\end{align}

\subparagraph{XLRUP format.}
The XLRUP format also marks BNN-specific steps with the keyword \textbf{b} and XOR-specific steps with \textbf{x}, with slight modifications in notation.
Specifically, the keyword (\textbf{i cb}) indicates a \texttt{clause-from-bnn} step in XLRUP.
For example, the following step records the derivation of the clause $(\neg x_1 \vee \neg x_3)$ with index 4, inferred from the first BNN constraint (Equation~\ref{eq:bnn-eg}) and the unit clause indexed by 3:
\begin{Verbatim}[fontsize=\small,commandchars=\\\{\}]
	\greenc{i cb} \yellowc{4} -1 -3 0 \yellowc{1} \greenc{u} \yellowc{3} 0
\end{Verbatim}

Similarly, the keyword (\textbf{i cx}) denotes a \texttt{clause-from-xor} step in XLRUP.
For instance, the following step records the derivation of the clause $(x_1 \vee x_2 \vee x_3)$ with index 9, inferred from the first XOR constraint (Equation~\ref{eq:xor-eg}):
\begin{Verbatim}[fontsize=\small,commandchars=\\\{\}]
	\greenc{i cx} \yellowc{9} 1 2 3 0 \yellowc{1} 0
\end{Verbatim}

\subparagraph{Derivation of UNSAT in Figure~\ref{fig:cnf-xor-bnn}.}
We use the FRAT-XOR-BNN proof as an example to illustrate the UNSAT derivation for the CNF-XOR-BNN formula shown in Figure~\ref{fig:cnf-xor-bnn}.
In the proof, after recording the original constraints using the keyword \textbf{o}, the first new clause $(\neg x_1 \vee \neg x_3)$ is derived by a \texttt{clause-from-bnn} step at index 4. This clause is inferred from the BNN constraint (Equation~\ref{eq:bnn-eg}) together with the unit clause $\neg x_4$ indexed by 3.
At this step, the solver makes a decision on $x_3$, which triggers propagation of the BNN constraint (Equation~\ref{eq:bnn-eg}) as described in Line~\ref{ln:prop-bnn-yf-prop} of Algorithm~\ref{alg:prop-bnn}.
Specifically, the assignments $\neg x_4$ and $x_3$ imply a {\false} assignment to the remaining literals in the BNN constraint.
Consequently, $\neg x_3$ (as the reason) and $\neg x_1$ (as the propagated assignment) are added to the reason clause at step 4.
Similarly, the reason clause $(x_2 \vee \neg x_3)$ is derived at step 5 from the same BNN constraint to infer a {\true} assignment to $x_2$.

Subsequently, $\neg x_1$ and $x_2$ lead to a conflict in the first input clause $(x_1 \vee \neg x_2)$, resulting in the unit learned clause $\neg x_3$ at step 6 through conflict analysis.
After backtracking to decision level 0 and applying unit propagation, the assignments $\neg x_1$ and $\neg x_2$ are derived at steps 7 and 8, respectively.
Finally, the assignments $\neg x_1, \neg x_2$, and $\neg x_3$ cause a conflict in the XOR constraint (Equation~\ref{eq:xor-eg}), from which the conflict clause $(x_1 \vee x_2 \vee x_3)$ is generated via a \texttt{clause-from-xor} step at step 9.
The empty clause is then derived at step 10, concluding the proof.
The XLRUP proof follows the same reasoning but uses a different syntax.

\subparagraph{Proof generation from {\cmsbnn}.}
{\cmsbnn} records \texttt{clause-from-bnn} steps in a lazy manner (\texttt{clause-from-xor} steps are handled similarly).
Specifically, the solver logs conflicts and unit propagations from BNN constraints during the search process and generates only the conflict clauses and relevant reason clauses during conflict analysis, which are recorded as \texttt{clause-from-bnn} steps.
Additionally, we track the list of unit clauses used to simplify BNN constraints during both pre-processing and in-processing, and incorporate them into \texttt{clause-from-bnn} steps following the keyword \textbf{u}.
We disable the variable replacement technique, which replaces a variable in a BNN constraint with another, as it requires reasoning over a BNN constraint and two clauses (or an XOR constraint).\footnote{
Such replacements can be logged using two clauses, $(x_1 \vee \neg x_2) \wedge (\neg x_1 \vee x_2)$, or an XOR constraint, $(x_1 \oplus x_2 = 0)$ to replace $x_1$ with $x_2$.
The clause-based replacement is compatible with clause steps, while the XOR-based replacement is compatible with XOR steps. However, neither is compatible with BNN-specific steps in our proof format.}
We also disable BNN (and XOR) propagation during the distillation phase~\cite{EB05}, as our purely clausal RUP proofs do not support BNN propagation.

\subparagraph{FRAT-XOR-BNN to XLRUP through {\fratxorbnn}.}
{\fratxorbnn} follows the \textit{lightweight} design principle of {\fratxor}~\cite{TYSM+24}: it does not verify the correctness of BNN-specific steps but delegates the checking of those steps to a formally verified tool, {\cakexlrup}.
Our primary modification involves tracking clauses derived from BNN constraints to ensure their correct usage in subsequent clausal steps and for the automatic elaboration of RUP proofs~\cite{BCH21}.

\subparagraph{Formally verified proof checking with {\cakexlrup}.}
Our {\cakexlrup} tool extends an earlier proof checker~\cite{TYSM+24} with efficient and verified BNN-specific reasoning.
The verification is customized to check \texttt{clause-from-bnn} (\textbf{i cb}) steps quickly.
Briefly speaking, the literals in each BNN constraint are stored in a bitset; whenever a clause $C$ is to be derived, the proof checker tracks propagations for all units from $\lnot{C}$ (and others provided in the proof hint) using the bitset, and accepts the derivation of $C$ if a contradiction is derived.
The procedure for checking conflicts is similar to Algorithm~\ref{alg:prop-bnn}.

A particularly useful optimization is to keep the size of each bitset minimal for the corresponding BNN constraint.
In straightforward numbering schemes for BNN neurons, the variables appearing in the resulting constraints are dense and contiguous, i.e., $x_i,x_{i+1},\dots,x_{i+n}$.
In such cases, the corresponding bitset will only allocate memory for the indices $i$ to $i+n$.
This optimization ensures that the overall representation takes $O(n)$ space where $n$ is the number of neurons in the BNN, as opposed to $O(nv)$, where $v$ is the total number of variables (if the bitset naively stored indices for every possible variable in the formula).

Note that, by design of the XLRUP extensions, our new BNN steps interact smoothly with the existing proof system for clauses and XOR constraints.
As with earlier versions, our extensions are formally verified down to machine code implementations using CakeML~\cite{TMKFON19CakeML}.

\subsection{Certified Counting for Quantitative BNN Reasoning}
\label{sec:quantitative-reasoning}
We present our certified model counter, {\ApproxMCCertBNN}, along with its certificate checker, {\CertCheckBNN}, designed for efficient quantitative reasoning on BNNs.

\subsubsection{Model Counting for CNF-BNN Formulas} \label{sec:bnn-counting}
Let us start by briefly explaining the key idea behind {\appmc}, a state-of-the-art approximate model counting algorithm and implementation~\cite{CMV13,YM23}.
It repeatedly samples and adds random XOR constraints to halve the solution space of the formula; eventually, after adding a number $m$ of such constraints, the number of remaining solutions $n$ will become sufficiently small.
Then, the overall solution count is approximated as $n\cdot2^m$.
The core requirement for an efficient implementation is the use of an incremental CNF-XOR solver to find solutions or determine that no further ones exist.

When the input formula $\varphi$ contains both clauses and BNN constraints---that is, when $\varphi$ is a CNF-BNN formula---the same counting algorithm can be used except that a CNF-XOR-BNN solver is required to compute the number of solutions to the combined CNF-XOR-BNN formula $\varphi\wedge XOR_1 \wedge \ldots \wedge XOR_m$.

To this end, we develop {\appmcbnn}, an approximate model counter built on top of our CNF-XOR-BNN solver, {\cmsbnn}, introduced earlier.
{\appmcbnn} takes a CNF-BNN formula as input and computes an approximate solution count.
The correctness of {\appmcbnn} is guaranteed by the following theorem;
a formalized proof in {\isabellehol} has established the correctness of {\appmc} for arbitrary Boolean theories~\cite{TY24,TYSM+24}.

\begin{theorem}
  \label{thm:cmsbnn-count}
	For a CNF-BNN formula $\varphi$, a tolerance parameter $\varepsilon$, and a confidence parameter $\delta$, {\appmcbnn} returns an approximate count $c$ such that
	\begin{align*}
		\Prob{\frac{|\sol|}{1+\varepsilon} \leq c \leq(1 + \varepsilon)|\sol|} \geq 1-\delta
	\end{align*}
\end{theorem}

We emphasize that {\appmcbnn} is the first model counter that natively supports BNN constraints. This enables it to use a succinct input representation (without extension variables), which significantly improves counting efficiency.

\subsubsection{Certifying CNF-BNN Counting} \label{sec:bnn-counting-cert}
The counting certification pipeline for CNF-BNN formulas follows the same framework as the certified CNF model counter, {\ApproxMCCert}, and its certificate checker, {\CertCheck}~\cite{TYSM+24}.
Specifically, the certified approximate CNF-BNN model counter, {\ApproxMCCertBNN}, takes a CNF-BNN formula as input and produces an approximate model count along with a certificate, which can be independently verified by the certificate checker, {\CertCheckBNN}, to ensure correct execution modulo randomness~\cite{TYSM+24}.

Most relevant for us, the design of certificates for {\CertCheck} records the number of XOR constraints used, as well as the solutions to the resulting CNF-XOR-BNN formulas, which are essential for estimating the model count.
During certificate checking, {\CertCheckBNN} verifies the correctness of the solutions to the CNF-XOR-BNN formulas and employs {\cmsbnn} and {\cakexlrup} to generate and validate UNSAT proofs, ensuring exhaustive enumeration of solutions.
If all solutions and UNSAT proofs pass verification by {\CertCheckBNN}, it outputs a certified model count; otherwise, an error is reported~\cite{TYSM+24}.

Our new {\CertCheckBNN} tool was built by extending the earlier formal verification.
Thanks to the genericity of the theorems proved in prior work~\cite{TY24}, the main verification task here was to develop a theory of CNF-BNN formulas in {\isabellehol} and then instantiate earlier results to derive a correct-by-construction certificate checking framework, including a formalized version of Theorem~\ref{thm:cmsbnn-count}.

\subsection{Bug Report from Certification} \label{sec:bnn-bug-report}
During the development of our certification pipeline, we identified and resolved a subtle bug in the implementation of {\cmsbnn}, related to the watching scheme for BNN constraints (Equation~\ref{eq:cond-card}).
As shown in Algorithm~\ref{alg:prop-bnn}, {\cmsbnn} maintains two counters--{\tcnt} (literals assigned {\true}) and {\ucnt} (unassigned literals)--to determine the propagation state of a BNN constraint.
Each literal is watched according to its polarity: assigning a positive literal increments {\tcnt} and decrements {\ucnt} (Lines~\ref{ln:prop-bnn-l-pos-begin}--\ref{ln:prop-bnn-l-pos-end}),
while assigning a negative literal decrements {\ucnt} (Line~\ref{ln:prop-bnn-l-neg}).

To preserve the AtLeastK form (Equation~\ref{eq:cond-card}) when the output literal $l_y$ is assigned {\false}, {\cmsbnn} flips the polarity of LHS literals.
However, the implementation mistakenly failed to update the polarity of watched literals during this process, resulting in incorrect BNN propagations.
This bug was nontrivial to detect through conventional testing but was successfully uncovered by our attempts at running benchmarks through the certification pipeline.%

%% file: sec/experiment.tex
\section{Experimental Evaluation} \label{sec:exp}
In this section, we evaluate the performance of our tools, namely, our proof-generating BNN-based solver ({\cmsbnn}), proof checker ({\cakexlrup}), certified BNN-based model counter ({\ApproxMCCertBNN}) and its certificate checker ({\CertCheckBNN}).
Experiments were conducted on a BNN robustness benchmark~\cite{BSSM+19}, which consists of 960 problem instances with varying BNN architecture sizes.

For comparison, we also transformed the BNN formulas into CNF and PB encodings and evaluated our tools against state-of-the-art certified solvers and counters.
Due to the large number of compared tools, we introduce them in their respective subsections.

\subparagraph{Setting.} All experiments were conducted on a high-performance computing cluster, where each node is equipped with an AMD EPYC-Milan processor featuring $2 \times 64$ physical cores and 512GB of RAM.
Each solver or counter was required to produce a checkable proof/certificate alongside its output.
For every tool, we set a time limit of 500 seconds for qualitative reasoning and 5,000 seconds for quantitative reasoning, with a memory limit of 16 GB.
For approximate counting, we used the default parameter values of $\delta=0.2$ and $\varepsilon=0.8$.

We report the PAR-2 score for each tool, a standard evaluation metric used in the SAT competition.
The PAR-2 score is the weighted average runtime across all instances, including the actual runtime for successfully completed instances, and double the time limit for instances that exceed the timeout (i.e., 1,000 seconds for qualitative reasoning and 10,000 seconds for quantitative reasoning in our setting).

Our experimental evaluation is designed to address the following research questions:
\begin{description}
	\item[(RQ1)] How does the certified solving performance of {\cmsbnn} compare to alternative certified approaches using CNF and PB encodings for BNNs?
	\item[(RQ2)] How does the certified counting performance of {\ApproxMCCertBNN} compare to the CNF-based baseline for BNNs?
\end{description}

\subparagraph{Summary.}
Overall, our approach significantly improves the runtime performance in both qualitative and quantitative reasoning for BNNs, and makes certified BNN verification practically feasible.
Specifically,
\begin{description}
	\item[(RQ1)] {\cmsbnn} and {\cakexlrup} produced fully certified answers for $99\%$ of the qualitative queries, achieving a $9\times$ speedup over alternative CNF and PB approaches, which fully certified only $62\%$ of all queries.
	\item[(RQ2)] {\ApproxMCCertBNN} and {\CertCheckBNN} fully certified results for $86\%$ of the quantitative queries, achieving a $218\times$ speedup over the CNF baseline, which could fully verify only $4\%$ of the queries.
\end{description}

\subsection{Certified Qualitative Reasoning for BNNs}
This section evaluates (RQ1) the runtime performance and proof-checking efficiency of {\cmsbnn} and {\cakexlrup} in answering qualitative queries for BNNs.
The baselines include the state-of-the-art CNF solver {\cadical}~\cite{DBLP:conf/cav/BiereFFFFP24} (f13d744) with its formally verified LRAT proof checker {\cakelpr}~\cite{DBLP:journals/sttt/TanHM23} (36b917a), and the PB solver {\roundingsat}~\cite{EN18} (73aaf09) with its proof checker {\veripb}~\cite{BGMN23} (178904a).\footnote{The {\veripb} toolchain also has a formally verified checker which we did not run for our evaluation. We opted for the static binary of {\veripb} from the SAT competition 2023, as our evaluation showed that the 2024 version performed slightly slower on our problem instances.}
We selected the 125 UNSAT instances (out of 960) as qualitative BNN verification tasks; certifying satisfiability of the remaining instances is straightforward so we reserve those for qualitative reasoning (counting).

A summary performance comparison is shown in Table~\ref{tab:solve-cert}.
From the top row, we observe that {\cmsbnn} successfully solved all 125 instances, achieving the lowest PAR-2 score of 16.
The lower row compares the overall performance of solving and proof checking across our three configurations.
All but one of the benchmarks were successfully certified by our approach; notably, although {\roundingsat} solved 123 instances, only 67 proofs were verified by {\veripb} within the time limit.
This shows the benefit of our native design for both solving and certification.

\begin{table}
	\caption{
	Runtime performance (time in seconds) for certified qualitative reasoning for BNNs over 125 UNSAT instances.
  The first row shows solver performance with proof generation, while the second row presents the combined performance of solving and proof checking for various toolchains.
	}
		\centering
    \begin{tabular}{l@{\hskip 0.1in} c@{\hskip 0.1in} c@{\hskip 0.1in}  c@{\hskip 0.3in} c@{\hskip 0.1in} c@{\hskip 0.1in} c@{\hskip 0.1in} }
			\toprule
			& {\cadical}	&  {\roundingsat}	& {\cmsbnn} \\
			\midrule
			Solved Instances   &   77		&     123     &    \textbf{125} \\
			PAR-2 Score 		&   478	&	 72    &    \textbf{16} \\
			\midrule
			& +{\cakelpr} & +{\veripb} & +{\cakexlrup} \\
			\midrule
			Solved Instances & 77 		&	67	 	&   \textbf{124}  \\
			PAR-2 Score 		 & 874    &	1,068 	& 	\textbf{60}  \\
			\bottomrule
\end{tabular}
\label{tab:solve-cert}
\end{table}

\begin{figure}[t]
	\centering
	\includegraphics[scale=0.85]{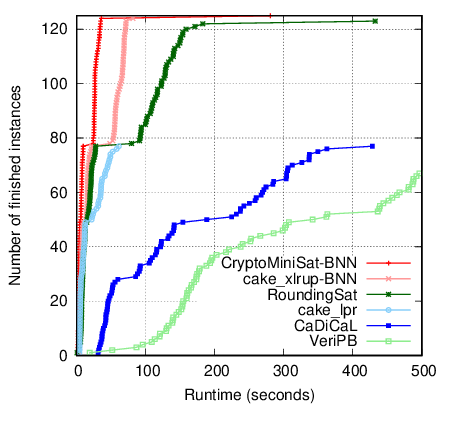}
  \includegraphics[scale=0.85]{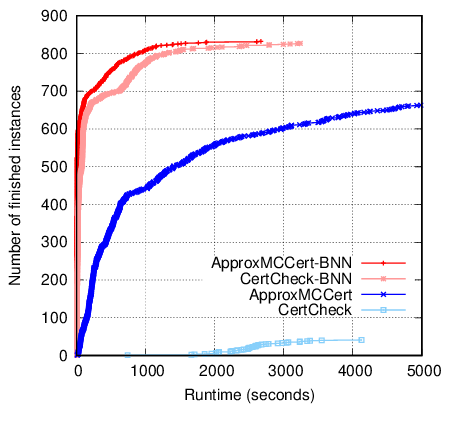}
	\caption{ (Left) Runtime performance comparison of solvers and proof checkers on qualitative reasoning benchmarks.
  (Right) Runtime performance comparison of counters and certificate checkers.
  Note that the solving/counting and checking times are plotted separately in both plots.
  }
	\label{fig:solve-check-time}
\end{figure}

Figure~\ref{fig:solve-check-time} (left) plots the cumulative number of solved or checked instances for a given time limit, i.e., each point $(x, y)$ indicates that the tool successfully completed $y$ instances within $x$ seconds.
{\cmsbnn}, together with {\cakexlrup}, consistently appears at the top of the figure, providing improved solving and checking performance compared to {\roundingsat} with {\veripb} and {\cadical} with {\cakelpr}.
For solving times, {\cmsbnn} is on average $26\times$ faster than {\cadical} and $3\times$ faster than {\roundingsat}.
In fact, {\cmsbnn} is also $2\times$ faster than {\minisatcs}~\cite{JR20}; however, the latter does not support UNSAT proof generation.
A detailed comparison with {\minisatcs} is presented in Appendix~\ref{sec:appendix-minisatcs-performance}.
For combined solving and checking times, {\cmsbnn+\cakexlrup} achieved a $9\times$ speedup over {\cadical+\cakelpr} and an $18\times$ speedup over {\roundingsat+\veripb}.
The ratios are calculated over the common solved (and checked) instances for each pair.

\subsection{Certified Quantitative Reasoning for BNNs}
This section evaluates (RQ2) the runtime performance and certificate-checking efficiency of {\ApproxMCCertBNN} and {\CertCheckBNN} for quantitative BNN queries.
We compare our approach against the certified CNF counter {\ApproxMCCert} and its certificate checker {\CertCheck}~\cite{TYSM+24}.
Additionally, we disabled the counting preprocessor {\arjun}~\cite{SM22} due to its poor performance on BNN benchmarks.
Our evaluation is conducted on the full 960 instances of the quantitative robustness benchmark for BNNs~\cite{BSSM+19}.

Table~\ref{tab:count-cert} presents the runtime performance of the counters and their corresponding certificate checkers.
In terms of counting performance (with certificate generation), we observed a significant improvement where {\ApproxMCCertBNN} solved 169 more instances and more than halved the PAR-2 score to 1,443.
For the combined performance of counting and certificate checking, {\ApproxMCCert} together with {\CertCheck} could produce fully certified results for only $4\%$ of the instances (41 out of 960) within the time limit.
In contrast our approach, {\ApproxMCCertBNN+\CertCheckBNN}, completed both counting and certificate checking for $86\%$ of the instances (827 out of 960), and substantially reduced the PAR-2 score from 19,256 to 3,068.

\begin{table}
	\caption{
	Runtime performance (time in seconds) for certified quantitative reasoning for BNNs over 960 counting instances.
  The first two columns show {\ApproxMCCert}'s counting performance, followed by the combined counting and certificate checking performance with {\CertCheck};
  the latter two columns similarly show the performance of {\ApproxMCCertBNN} together with {\CertCheckBNN}.}
	\centering
	\begin{tabular}{l@{\hskip 0.1in} c@{\hskip 0.1in}  c@{\hskip 0.3in} c@{\hskip 0.1in} c@{\hskip 0.1in}}
		\toprule
		&  {\ApproxMCCert}	& +{\CertCheck} 	& {\ApproxMCCertBNN} 			& +{\CertCheckBNN} \\
		\midrule
		Finished       		&     663   &  41			  &  \textbf{832} 	&    \textbf{827} \\
		PAR-2 Score 		&     3,778   & 19,256   &  \textbf{1,443}	  		& 	 \textbf{3,068} \\
		\bottomrule
	\end{tabular}
	\label{tab:count-cert}
\end{table}

Figure~\ref{fig:solve-check-time} (right) shows the number of instances completed by the counters and checkers over time.
Here, {\ApproxMCCertBNN} consistently solved the largest number of instances and it is closely trailed by {\CertCheckBNN}---thus, {\ApproxMCCertBNN} offers superior counting runtime and its generated certificates were also readily checked by {\CertCheckBNN}.
In terms of average counting and certification times, {\ApproxMCCertBNN+\CertCheckBNN} achieved a $218\times$ speedup over {\ApproxMCCert+\CertCheck} on their common, fully certified instances.

Upon deeper investigation, a key issue for the certificates of {\CertCheck} is that the CNF encodings of BNN benchmarks lead to \emph{projected} counting instances with an extremely large number of extension variables.
The counting certificate format must record satisfying assignments for all variables, which leads to substantial file size, memory, and timing overheads in the CNF-based approach.

In fact, {\ApproxMCCertBNN} outperforms even existing counters that do not provide certification for quantitative reasoning on BNNs, such as the exact CNF counter {\Ganak}~\cite{SRSM19}, the exact PB counters {\PBCount}~\cite{YM24}, and the approximate PB counter {\ApproxMCPB}~\cite{YM21}.
A detailed comparison is provided in Appendix~\ref{sec:appendix-counting-performance}.

%% file: sec/conclusion.tex
\section{Conclusion} \label{sec:conclusion}
Certified verification is critical for neural networks deployed in safety-critical applications.
This work presents an efficient certified solver for qualitative reasoning and a certified approximate model counter for quantitative reasoning on BNNs.
Our approach significantly outperforms existing CNF and PB-based methods and produces fully certified results for the vast majority of queries, making certified BNN verification practically feasible.
Looking forward, this framework opens the door to certifying large-scale binarized vision and language models~\cite{HLZL+23,ZGCL+23},
as well as extending certification to quantized neural networks for efficient on-device deployment~\cite{LTTY+24,XLSW+23}.

%% file: sec/appendix.tex
\section{Solving Performance Comparison with {\minisatcs}}
\label{sec:appendix-minisatcs-performance}
We present a performance comparison between {\cmsbnn} and {\minisatcs}~\cite{JR20}, excluding proof generation, which is not supported by {\minisatcs}.
Our evaluation is conducted on the full set of 960 instances from the quantitative robustness benchmark for BNNs~\cite{BSSM+19}, including both SAT and UNSAT cases.
We set a time limit of 500 seconds and a memory limit of 4GB for each instance. 

\begin{figure}[t]
	\centering
	\includegraphics[scale=0.85]{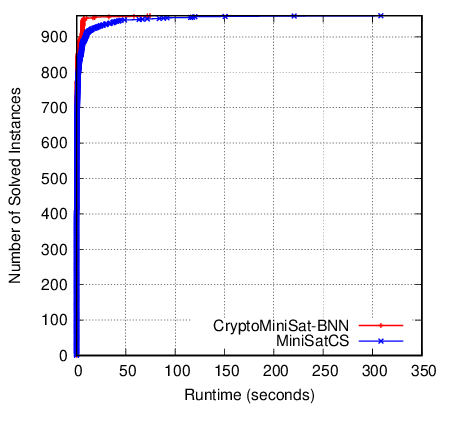}
	\caption{Runtime performance comparison of {\cmsbnn} and {\minisatcs}.}
	\label{fig:minisatcs-time}
\end{figure}

As shown in Table~\ref{tab:minisatcs}, both {\cmsbnn} and {\minisatcs} solved all instances. However, {\cmsbnn} achieved a lower PAR-2 score of 1, compared to 3 for {\minisatcs}.
Figure~\ref{fig:minisatcs-time} illustrates the number of solved instances over time.
{\cmsbnn} solved all instances within 75 seconds, whereas {\minisatcs} required up to 309 seconds.
Overall, {\cmsbnn} achieved a $2\times$ speedup over {\minisatcs}.

\begin{table}
	\caption{Runtime performance comparison of {\cmsbnn} and {\minisatcs} over 960 instances.}
	\centering
	\begin{tabular}{l@{\hskip 0.1in} c@{\hskip 0.1in} c@{\hskip 0.1in}}
		\toprule
		& {\minisatcs}	&  {\cmsbnn}  \\
		\midrule
		Solved Instances   &   960		&     960    \\
		PAR-2 Score 		&   3	&	 1   \\
		\bottomrule
	\end{tabular}
	\label{tab:minisatcs}
\end{table}

\section{Counting Performance Comparison}
\label{sec:appendix-counting-performance}
We present the counting performance comparison between {\ApproxMCBNN} and state-of-the-art CNF counters, {\Ganak}~\cite{SRSM19} and {\ApproxMC}~\cite{YM23}, and PB counters, {\PBCount}~\cite{YM24} and {\ApproxMCPB}~\cite{YM21} in Table~\ref{tab:count}; {\ApproxMCBNN} refers to {\ApproxMCCertBNN} without certification generation.
We also repeat the numbers for  {\ApproxMCCertBNN} for comparison.
Each counter uses a memory limit of 4GB.

The CNF counters demonstrated the worst performance. The exact and approximate counter, {\Ganak} and {\ApproxMC} solved only 424 and 770 out of 960 instances, respectively, with the two highest two PAR-2 scores in the table.
The approximate PB counter, {\ApproxMCPB} solved 816 instances and lowered the PAR-2 score to 1,741.
{\PBCount} failed to solve any instance so we removed it from the table.
Lastly, {\ApproxMCBNN} achieved the best performance with 868 instances solved and further reduced the PAR-2 score to 1,128. {\ApproxMCBNN} solved 98 more instances than the best CNF counter (\ApproxMC) and outperformed the best PB counter {\ApproxMCPB} by 52 more solved instances.
Even with the overhead of certificate generation, {\ApproxMCCertBNN} still outperforms other non-BNN-native approaches.

\begin{table}
	\caption{Counting performance comparison over 960 instances.}
	\centering
	\begin{tabular}{l@{\hskip 0.1in} c@{\hskip 0.1in} c@{\hskip 0.1in}  c@{\hskip 0.1in} c@{\hskip 0.1in} c@{\hskip 0.1in}}
		\toprule
		& {\Ganak}	&  {\ApproxMC}	& {\ApproxMCPB} & {\ApproxMCBNN} & {\ApproxMCCertBNN} \\
		\midrule
		Counted Instances   &   424		&     770     &    816 			&  \textbf{868}   &  832 	  \\
		PAR-2 Score 		&   6248	&	 2600    &    1741 		        & \textbf{1128}  &  1,443 \\
		\bottomrule
	\end{tabular}
	\label{tab:count}
\end{table}

\begin{figure}[t]
	\centering
	\includegraphics[scale=0.85]{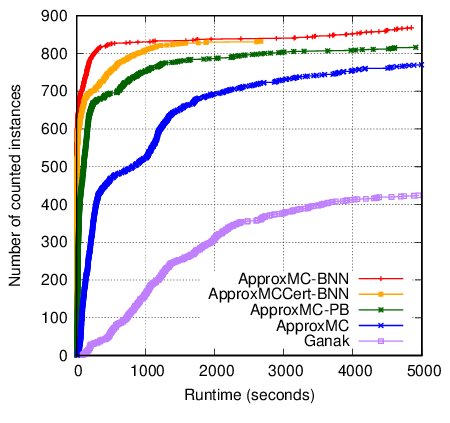}
	\caption{Runtime performance comparison between counters.}
	\label{fig:count-time}
\end{figure}

Figure~\ref{fig:count-time} compares the counting performance in terms of counted instances per runtime.
The plot shows that {\ApproxMCBNN} consistently solves the most instances at any time limit, followed by {\ApproxMCPB}, {\ApproxMC}, and {\Ganak}.